\documentclass[journa,onecolumn]{IEEEtran}
\ifCLASSINFOpdf
\else
\fi
\usepackage{graphicx}
\usepackage{comment}
\usepackage{cite}
\usepackage[fleqn]{amsmath}
\usepackage{url}

\begin{document}
%
\title{Land Cover Classification via Multi-temporal Spatial Data by Recurrent Neural Networks}
%
%
%

\author{Dino Ienco,
        Raffaele Gaetano,
        Claire Dupaquier
        and Pierre Maurel
\thanks{D. Ienco is with UMR-TETIS laboratory, IRSTEA, Montpellier, France and with LIRMM laboratory, Montpellier, France (email: dino.ienco@irstea.fr).}%
\thanks{C. Dupaquier and P. Maurel are with UMR-TETIS laboratory, IRSTEA, Montpellier, France (email: claire.dupaquier@irstea.fr, pierre.maurel@irstea.fr).}
\thanks{R. Gaetano is with UMR-TETIS laboratory, CIRAD, Montpellier, France (email: raffaele.gaetano@cirad.fr).}
}

\maketitle

\begin{abstract}
Nowadays, modern earth observation programs produce huge volumes of satellite images time series (SITS) that can be useful to monitor geographical areas through time. How to efficiently analyze such kind of information is still an open question in the remote sensing field. Recently, deep learning methods proved suitable to deal with remote sensing data mainly for scene classification (i.e. Convolutional Neural Networks - CNNs - on single images) while only very few studies exist involving temporal deep learning approaches (i.e Recurrent Neural Networks - RNNs) to deal with remote sensing time series.

In this letter we evaluate the ability of Recurrent Neural Networks, in particular the Long-Short Term Memory (LSTM) model, to perform land cover classification considering multi-temporal spatial data derived from a time series of satellite images. We carried out experiments on two different datasets considering both pixel-based and object-based classification. The obtained results show that Recurrent Neural Networks are competitive compared to state-of-the-art classifiers, and may outperform classical approaches in presence of low represented and/or highly mixed classes. We also show that using the alternative feature representation generated by LSTM can improve the performances of standard classifiers.

\end{abstract}


%
\IEEEpeerreviewmaketitle

\section{Introduction}

\IEEEPARstart{M}{odern} earth observation programs produce huge volumes of remotely sensed data every day. Such information can be organized in time series of satellite images that can be useful to monitor geographical zones through time. Efficiently manage and analyze remote sensing time series is still an open challenge in the remote sensing field~\cite{Karpatne16}.

In the context of land cover classification, exploiting time series of satellite images, instead that one single image, can be fruitful to distinguish among classes based on the fact they have different temporal profiles~\cite{Abade15}. Despite the usefulness of temporal trends that can be derived from remote sensing time series, most of the proposed strategies~\cite{Flamary15,Heine16} directly apply standard machine learning approaches (i.e. Random Forest, SVM) on the stacked images. Since these approaches did not model temporal correlations, they manage features independently from each others, ignoring any temporal dependency which data may exhibit.
Recently, the deep learning revolution~\cite{Zhang16} has shown that neural network models are well adapted tools to manage and automatically classify remote sensing data. While standard CNNs techniques are well suited to deal with spatial autocorrelation, the same approaches are not adapted to correctly manage long and complex temporal dependencies~\cite{BengioCV13}. A family of deep learning methods especially tailored to cope with temporal correlations are Recurrent Neural Networks~\cite{BengioCV13} and, in particular, Long Short Term Memory (LSTM) networks~\cite{GreffSKSS15}. Such models explicitly capture temporal correlations by recursion and they have already proved to be effective in different domains such as 
speech recognition~\cite{Graves13}, natural language processing~\cite{LinzenDG16}, image completion~\cite{OordKEKVG16}.
Only recently, in the remote sensing field, the work proposed in~\cite{Lyu16} performs preliminary experiments with LSTM model on a (small) time series composed of only two dates to perform supervised change detection. The task was modeled as a binary classification problem (change vs. no-change). To the best of our knowledge, RNNs (i.e. LSTM) have not yet been considered to deal with land cover classification of deeper time series. 
Like any other deep learning model~\cite{BengioCV13}, LSTM can be used as a classifier itself or employed to extract new discriminative features (or representation). In the latter case the extracted features are successively used to feed a standard learning algorithm that does not consider temporal dependencies (i.e. Random Forest, Naive Bayes, KNN, SVMs).

In this letter we evaluate the quality of RNN models - Long Short Term Memory - to deal with land cover classification via multi-temporal spatial data that are derived from SITS. More in detail, we perform experiments on two study areas: i) the \textit{THAU} basin, a site located in the south of France, from which we obtain a time series of 3 dates, and ii) the \textit{REUNION ISLAND}, a region of France located in the Indian Ocean (east of Madagascar) from which we derive a 23-date times series. 
While on the first area we conduct an object-oriented classification, on the second site we have performed a pixel-based prediction showing the general applicability of RNNs models to both object and pixel-level analysis. We also assess the RNN model (i.e. LSTM) as feature extractor evaluating the quality of the new generated features to feed the same baseline classifiers we have used as reference methods.

The rest of the paper is organized as follows: Section~\ref{sec:method} introduces the LSTM unit and specifies the network architecture we propose, Section~\ref{sec:data} describes the two datasets, the time series characteristics and the preprocessing we have performed on the source data. Experimental setting and results are discussed in Section~\ref{sec:expe}. Conclusions are drawn in Section~\ref{sec:conclu}.

%
%
%
%


\section{Method}
\label{sec:method}
We propose a neural architecture involving LSTM unit to deal with land cover classification via multi-temporal spatial data. We also assess the representation learned by the RNN model with standard classification strategies commonly used to perform prediction in the remote sensing field.

\subsection{Long-Short Term Memory}
\begin{figure}[ht!]
\centering
\includegraphics[height=0.1\textwidth,width=0.30
\textwidth]{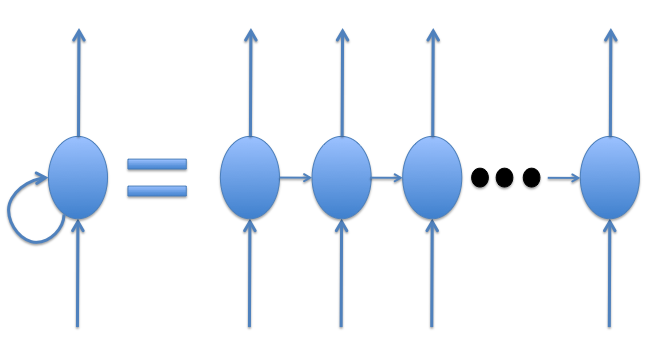}
\caption{Recurrent Neural Network Unit (on the left) and unfolded structure (on the right). \label{fig:RNN}}
\end{figure}

Recurrent Neural Networks are well established machine learning techniques that demonstrate their quality in different domains such as speech recognition~\cite{Graves13}, signal processing~\cite{SomaMSFN15}, natural language processing~\cite{LinzenDG16} and image completion~\cite{OordKEKVG16}. Differently from standard feed forward networks (i.e. CNNs), RNNs explicitly manage temporal data dependencies since the output of the neuron at time t-1 is used, together with the next input, to feed the neuron itself at time t. A sketch of a typical RNN neuron is depicted in Figure~\ref{fig:RNN}.

\begin{figure}[ht!]
\centering
\includegraphics[width=0.50
\textwidth]{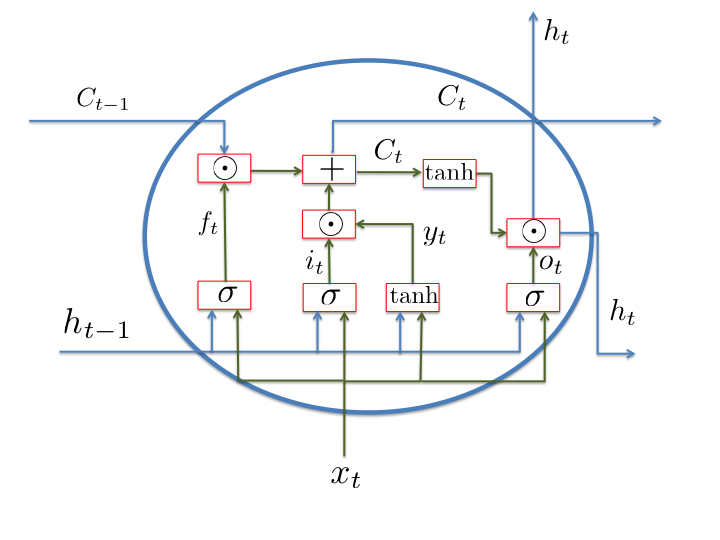}
\caption{The structure of the LSTM unit in which arrows indicate directed connection. Blue lines show the direction in which the information will flow while green lines underline internal flows. Red rectangles represent operations to combine or transform the different information. \label{fig:LSTM}}
\end{figure}

The most well-known type of RNN is the Long-Short Term Memory (LSTM)~\cite{GreffSKSS15} model. There are many variants of LSTM network~\cite{GreffSKSS15} but here we refer to the architecture proposed in~\cite{GersSC00}.
LSTM models were mainly introduced with the purpose to learn long term dependencies~\cite{GreffSKSS15}, since previous RNN models failed in this task due to the problem of vanishing and exploding gradients. The equations (1), (2), (3), (4), (5) and (6) formally describes the LSTM neuron while Figure~\ref{fig:LSTM} graphically depicts the LSTM unit. The symbol $\odot$ indicates an element-wise multiplication while $\sigma$ and $\tanh$ represent Sigmoid and Hyperbolic Tangent respectively.
The input of the LSTM is a sequence of variables ($x_1$, ..., $x_N$) where a generic element $x_t$ is a feature vector and $t$ refers to the corresponding timestamp. RNN models are able to manage variable-length data sequences.

The LSTM unit is composed of two cell states, the memory $C_t$ and the hidden state $h_t$, and three different gates, the input gate $i_t$, the forget gate $f_t$ and the output gate $o_t$ that are employed to control the flow of information. All the three gates combine the current input $x_t$ with the hidden state $h_{t-1}$ coming from the previous timestamp.
The gates have also two important functions: i) they regulate how much information have to be forgotten/reminded during the process; ii) they deal with the problem of vanishing/exploding gradients. We can observe that the gates are implemented by a sigmoid. This function returns values between 0 and 1 where, in this context, 0 indicates that the information is completely forgotten and 1 means that the information is completely retained.

The LSTM unit uses also a temporary cell state $y_t$ that rescales the current input always taking into account the previous hidden state. This temporary cell is implemented by an hyperbolic tangent function that returns values between -1 and 1. Both sigmoid and hyperbolic tangent are applied element-wise.

The input gate $i_t$ regulates how much of the current information needs to be maintained ($i_t \odot y_t$) while the forget gate $f_t$ indicates how much of the previous memory needs to be retained at the current step ($f_t \odot c_{t-1}$). Finally, the output gate impacts on the new hidden state $h_t$ deciding how much information of the current memory will be outputted to the next step. The different $W_{**}$ matrices and bias coefficients $b_{*}$ are the parameters learned during the training of the model. Both, the memory $C_t$ and the hidden state $h_t$ are forwarded to the next time step. 

\begin{align}
i_{t} = \sigma(W_{ix} x_{t} + W_{ih} h_{t-1} + b_i  ) \tag{1}\\
f_{t} = \sigma(W_{fx} x_{t} + W_{fh} h_{t-1} + b_f  ) \tag{2}\\
y_{t} = \tanh(W_{yx} x_{t} + W_{yh} h_{t-1} + b_y  )  \tag{3}\\
c_t = i_t \odot y_t + f_t \odot c_{t-1}  \tag{4}\\
o_{t} = \sigma(W_{ox} x_{t} + W_{oh} h_{t-1} + b_o  ) \tag{5}\\
h_t = o_t \odot \tanh(c_t) \tag{6}
\end{align}

\subsection{LSTM-Based Time Series Classification}
The LSTM neuron learns an internal representation of the input sequences (in our case objects or pixels time series) but it does not make any prediction by itself. To perform the classification task, we stack on top of the LSTM neuron a SoftMax layer~\cite{Graves13} to accomplish the final multi-class prediction. The SoftMax layer has as many neurons as the number of the classes to predict. We choose SoftMax instead of Sigmoid function because the value of the SoftMax layer can be seen as a probability distribution over the classes that sum to 1 while each of the Sigmoid neurons can output a value between 0 and 1. This is due to the fact that, for the SoftMax neuron, the values are normalized per layer while no normalization is performed in the case of Sigmoid layer. This is why, in our context (multi-class prediction), we prefer the SoftMax instead of Sigmoid layer since we know that our samples exclusively belong to a single class.
From an architectural point of view, the connection between the LSTM and the SoftMax layer is realized fully connecting the last hidden state vector produced by the LSTM unit with the SoftMax neurons. 

\subsection{Representation Learning with LSTM for time series data}
\label{sec:RLLSTM}
Standard deep learning approaches can also be seen as a way to produce a new, more discriminative representation of the original data~\cite{BengioCV13}. Another way to assess the quality of LSTM unit for our land cover classification task is to use the features learned by the LSTM layer to feed a standard classifier. More in detail, we propose to employ the last hidden state vector, produced by the LSTM unit, as new data representation and, successively, train standard machine learning classifiers over such new set of features.


\section{Data}
\label{sec:data}

In order to prove the generality of our proposal, it has been tested over two different remote-sensing based datasets. The first is a collection of spatial objects described by a set regional statistics extracted from very high spatial resolution imagery (VHSR), but with a limited temporal depth. The second one is a pixel-based dataset, more noisy but richer in both spectral and temporal resolution. Detailed descriptions are provided in the following subsections.

\subsection{THAU dataset}
The first dataset has been generated using a time series of Pl\'{e}iades VHSR images (2~m) acquired in the context of the Airbus DS/Spot Image distribution (July and September 2012, March 2013, $\copyright$ CNES). The study site is the \textit{THAU} Basin located in the South of France, close to Montpellier. It covers an area of 42\,000 ha with 70\% of land area. The north is mainly composed of agricultural fields (i.e. vineyards) and natural spaces while the south is dominated by urban and industrial zones. For each date, two orthorectified, atmospherically corrected scenes are mosaicked. 

Using the multi-temporal stack, a segmentation has been performed to extract a consistent multi-temporal object layer. Segmentation was performed using the Multiresolution Segmentation technique~\cite{Baatz2000} available in the \textit{eCognition Developer} software. Each object has been then featured using statistical mean and standard deviation using the four native bands (blue, green, red and near-infrared) and the NDVI. A total of 10 features are computed per object and per date (5 means and 5 standard deviations).

The so obtained segments have been subsequently filtered and labeled in 11 different classes by visual inspection. A total of 15\,196 objects is retained. The set of classes with the relative cardinality is reported in Table~\ref{tab:occSolDistrib}.

\begin{table}[ht!]
\centering
\scriptsize
\begin{tabular}{|l|c|c|} \hline
\textbf{ID} & \textbf{Land Cover Class} & \textbf{N. of Objects} \\ \hline
(1) & Tree crops & 600 \\ \hline
(2) & Forests and woods & 2\,445 \\ \hline
(3) & Water & 556 \\ \hline
(4) & Summer crops & 81 \\ \hline
(5) & Winter crops & 677 \\ \hline
(6) & Grasslands & 3\,882 \\ \hline
(7) & Sclerophyll vegetation  & 2\,457 \\ \hline
(8) & Truck farming & 227 \\ \hline 
(9) & Bare soils & 299 \\ \hline
(10) & Salt marshes & 236 \\ \hline
(11) & Vineyards & 3\,735 \\ \hline
\end{tabular}
\caption{Land Cover Classes and their cardinality for the \textit{THAU} time series dataset \label{tab:occSolDistrib}}
\end{table}

\subsection{REUNION ISLAND dataset}
The second dataset has been generated from an annual time series of 23 Landsat 8 images acquired in 2014 above the Reunion Island (2866 $\times$ 2633 pixels at 30~m spatial resolution), provided at level 2A\footnote{The source data are provided by the French \textit{P\^ole Th\'ematique Surfaces Continentales THEIA} (\url{www.theia-land.fr}) and preprocessed by the \textit{Multi-sensor Atmospheric Correction and Cloud Screening} (MACCS) level 2A processor~\cite{Hagolle2015} developed at the French National Space Agency (CNES) to provide accurate atmospheric, environmental and geometric corrections as well as precise cloud masks.}. Source data have been further processed to fill cloudy observations via pixel-wise multi-temporal linear interpolation on each multi-spectral band (OLI) independently, and compute complementary radiometric indices (NDVI, NDWI and brightness index - BI). A total of 10 features (7 surface reflectances plus 3 indices) are considered for each pixel at each timestamp.

Reference land cover data has been built using two publicly available dataset, namely the 2012 \textit{Corine Land Cover} (CLC) map and the 2014 farmers' graphical land parcel registration (\textit{R\'egistre Parcellaire Graphique} - RPG). The most significant classes for the study area have been retained, and a spatial processing (aided by photo-interpretation) has also been performed to ensure consistency with image geometry. Finally, a pixel-based random sampling of this dataset has been applied to provide an almost balanced ground truth. The final reference dataset consists of a total of 37\,900 pixels distributed over 9 classes as reported in Table~\ref{tab:occSolDistribR}.

\begin{table}[ht!]
\centering
\scriptsize
\begin{tabular}{|l|c|c|} \hline
\textbf{ID} & \textbf{Land Cover Class} & {N. of Pixels} \\ \hline
(1) & Urban areas & 10\,000 \\ \hline
(2) & Other built-up surfaces & 1\,500 \\ \hline
(3) & Forests & 10\,000 \\ \hline
(4) & Sparse Vegetation & 5\,095 \\ \hline
(5) & Rocks and bare soil & 3\,729 \\ \hline
(6) & Grassland & 1\,744 \\ \hline
(7) & Sugarcane crops & 2\,832 \\ \hline
(8) & Other crops & 1\,500 \\ \hline 
(9) & Water & 1\,500 \\ \hline
\end{tabular}
\caption{REUNION ISLAND}
\label{tab:occSolDistribR}
\end{table}

\section{Experimental Results}
\label{sec:expe}
In this section we report the experimental settings and we discuss the results we obtained on the two SITS datasets we presented in Section~\ref{sec:data}.

\subsection{Experimental Settings}
We compare the LSTM-based Time Series Classification model to standard machine learning approaches commonly employed to perform land cover classification from multi-temporal spatial data~\cite{Flamary15,Heine16}. We also assess the value of the representation learned by the proposed model following the idea described in Section~\ref{sec:RLLSTM}.

To our purpose, we use Random Forest (RF) and Support Vector Machine (SVM) as standard classification strategies.
For the \textit{RF} model, we set the number of generated trees equals to 400 and we allow a maximum tree depth of 10. For the \textit{SVM} model we use RBF kernel with complexity parameter and gamma equal to 100 and 0.01 respectively. For Random Forest we used the python implementation supplied by the Scikit-learn library~\cite{scikit-learn} while for \textit{SVM} we use the LibSVM implementation~\cite{CC01a}. The same \textit{RF} and \textit{SVM} settings are used for both original data and the new representation learned by the LSTM-Based Time Series Classification model.
For the latter, we set the number of hidden dimensions equal to 512, an initial learning rate equals to $5 \times 10^{-4}$ and a decay of $5 \times 10^{-5}$. We implement the model via the \textit{Keras} python library~\cite{chollet2015} with \textit{Theano} as back end. We used as optimization method the \textit{RMSprop} strategy that is commonly employed to train LSTM units~\cite{DauphinVCB15}.  The model is trained for 200 epochs with a batch size equals to 20.
We named \textit{RF(LSTM)} (resp. \textit{SVM(LSTM)}) the Random Forest (resp. \textit{SVM}) learned over the new feature space induced by our RNN model. More in detail, each training and test instance is transformed in a 512 feature vector (the dimension of the hidden state of the LSTM neuron) and, successively, the classifiers are learned from this new representation instead of the original data.  

To validate the different methods, we perform a 5-fold cross validation. Due to the unbalanced nature of the two time series datasets, in order to assess classification performances we use not only the Global Accuracy and Kappa measures, but we also provide average and per-class F-Measure.

\subsection{Results and Discussions}
\begin{figure}[ht!]
\centering
\includegraphics[width=0.95\columnwidth]{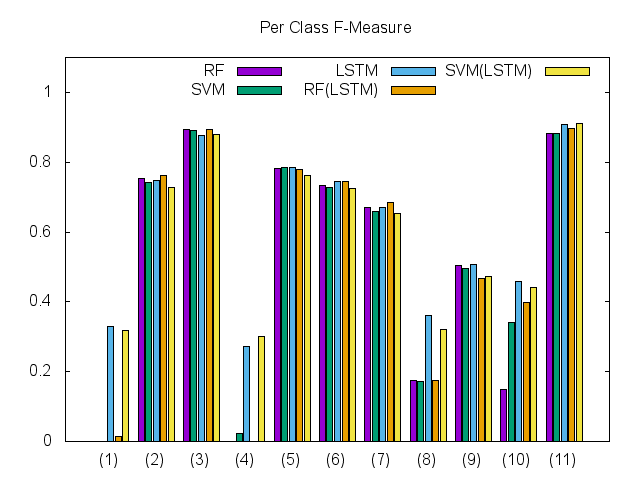}
\caption{Per Class F-Measure of the different approaches on the THAU dataset. \label{fig:perClassThau}}
\end{figure}

\begin{figure}[ht!]
\centering
\includegraphics[width=0.95
\columnwidth]{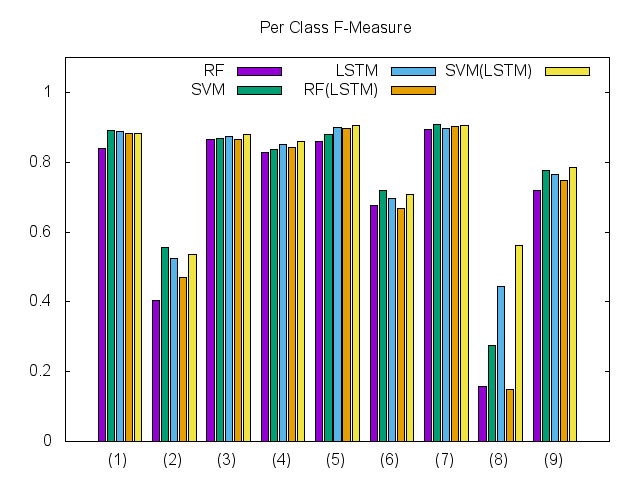}
\caption{Per Class F-Measure of the different approaches on the REUNION ISLAND dataset. \label{fig:perClassRI}}
\end{figure}

The Tables~\ref{tab:overalClassificationT} and~\ref{tab:overallClassificationRI} and Figures~\ref{fig:perClassThau} and~\ref{fig:perClassRI} summarize the results we have obtained on the two SITS datasets.

Considering the \textit{THAU} dataset, Table~\ref{tab:overalClassificationT} depicts the average values of \textit{Accuracy}, \textit{F-Measure} and \textit{Kappa} for the different methods. We can observe that the \textit{LSTM}-based classifier outperforms both \textit{RF} and \textit{SVM} approaches regarding all the three metrics. The more important gain is reached when the average \textit{F-Measure} is taken into account, the \textit{LSTM}-based classifier obtaining a score of 74.63\% while the second best method (\textit{SVM(LSTM))} attains a score equals to 73.31\%.  Interestingly, we can also highlight that, for the \textit{THAU} dataset, the classifiers trained on the features (representation) learned by the Recurrent Neural Network \textit{RF(LSTM)} and \textit{SVM(LSTM)} exhibit better performances than the same classifiers coupled with the original time series data when the \textit{F-Measure} is considered. In terms of Accuracy, the behavior is comparable considering the \textit{SVM} vs \textit{SVM(LSTM)} while the \textit{RF} model clearly benefit of the new data representation.

A more detailed assessment is provided in Figure~\ref{fig:perClassThau} where the per-class \textit{F-Measure} is reported. The first point we can highlight is that, for classes with few reference samples (i.e. (1),(4),(8), (9) and (10)), the \textit{LSTM} network neatly outperforms standard approaches, which in some cases (i.e. (1) and (4)) completely miss the classes, while it obtains similar or slightly better results w.r.t. \textit{RF} and \textit{SVM} for well represented classes. The second point is related to the comparison between standard approaches trained on the original data and the same methods powered by features learned by our proposal. We can see that the use of the new learned representation improves the performances of both \textit{RF} and \textit{SVM}. Again, this fact is particularly evident on critical classes like (1), (4), (8) (for SVM) and (10) (for RF).

\begin{table}[ht!]
\centering
\scriptsize
\begin{tabular}{|l|c|c|c|}\hline
Method & Accuracy & F-Measure & Kappa \\ \hline \hline
RF & 74.20\% & 71.58\% & 0.68 \\ \hline
SVM & 73.66\% & 71.35\% &  0.67\\ \hline
LSTM & 75.15\% & \textbf{74.63}\% & \textbf{0.69} \\ \hline
RF(LSTM) & \textbf{75.34}\% & 72.95\% & \textbf{0.69}\\ \hline
SVM(LSTM) & 73.50\% & 73.31\% & 0.67\\ \hline
\end{tabular}
\caption{5-Fold Cross Validation results \label{tab:overalClassificationT} on the \textit{THAU} dataset}
\end{table}

Table~\ref{tab:overallClassificationRI} summarizes the results on the \textit{REUNION ISLAND} time series dataset. Similarly to the previous case, the \textit{LSTM}-based classifier behaves better than the standard machine learning methods considering \textit{Accuracy} and \textit{F-Measure}; still in accord with the previous results, also the classifiers trained on the new features, obtained by the proposed network, outperform their counterparts trained on the original feature space. Conversely to the previous experiment, in this case the highest value of \textit{F-Measure} is reached by the \textit{SVM(LSTM)} but we can observe that the \textit{LSTM}-based classifier still reaches competitive results. Figure~\ref{fig:perClassRI} reports the per-class \textit{F-Measure} results on the \textit{REUNION ISLAND} dataset. Also in this case we can note that the \textit{LSTM}-based approach works well for low represented and difficult classes (i.e. (8) - "Other Crops") and it remains competitive on all the other classes.

As expected, the combined optimization and learning of a new feature representation along with classification, proper to all deep learning approaches, provides here a better support to discriminate among the different classes. In addition, all these results indicate that the \textit{LSTM} model is well suited to capture long-short temporal dependencies as opposed to common classification approaches where 
all the information are managed at the same level forgetting temporal correlations. This is particularly evident on low represented and highly mixed classes: \textit{Tree Crops}, \textit{Summer crops} and \textit{Truck Farming} (resp. \textit{Other Crops}) for the \textit{THAU} (resp. \textit{REUNION ISLAND}) dataset. 
All these classes are related to agricultural activities whose time patterns are strongly varying due to the heterogeneity of practices, which make the corresponding classes detectable only considering short portions of the time series in which crop conditions are comparable.


We remind that our proposal uses only one \textit{LSTM} layer while more layers can be stacked together to build more complex architectures~\cite{Graves13}. This is out of our scope and we leave this point for future researches, since our main objective is to highlight the quality and the suitableness of RNNs methods to manage and analyze SITS data.

\begin{table}[ht!]
\centering
\scriptsize
\begin{tabular}{|l|c|c|c|}\hline
Method & Accuracy & F-Measure & Kappa \\ \hline \hline
RF & 81.19\% & 79.41\% &   0.77\\ \hline
SVM & 83.82\% & 82.74\% & 0.80 \\ \hline
LSTM & 83.98\% & 83.56\% & 0.80 \\ \hline
RF(LSTM) & 83.12\% & 81.49\% & 0.79\\ \hline
SVM(LSTM) & \textbf{84.61}\% & \textbf{84.41}\% & \textbf{0.81}\\ \hline
\end{tabular}
\caption{5-Fold Cross Validation results \label{tab:overallClassificationRI} on the REUNION ISLAND dataset}
\end{table}

\section{Conclusion}
\label{sec:conclu}
In this letter we asses the benefit of using Recurrent Neural Network (LSTM) to perform land cover classification via multi-temporal spatial data. We have validated the proposed model on two different SITS based datasets showing that the proposed framework efficiently deals with both pixel-based and object-based classification. 

The proposed framework proved competitive, yet outperforming compared to classical approaches, with the remarkable advantage of improving the quality of the predictions on ``weak'' classes from unbalanced datasets. We also highlight that the proposed \textit{LSTM}-based classification model can be used as feature extractor to learn a new data representation that positively affect the performances of standard classification approaches on SITS data.


%

\section*{Acknowledgment}

The authors acknowledge also the National Research Agency in the framework of the program "Investissements d'Avenir" for the GEOSUD project (ANR-10-EQPX-20) for the distribution of the Pl\'{e}iades satellite images.

\ifCLASSOPTIONcaptionsoff
  \newpage
\fi



%
\bibliographystyle{plain}
\bibliography{paper.bib}


%








\end{document}